\documentclass{amia}
\usepackage{lipsum} 

\usepackage{algorithm}
\usepackage{algorithmic}
\usepackage{amsmath}
\usepackage{multirow}
\usepackage{hyperref}
\usepackage[dvipsnames]{xcolor}
\usepackage{booktabs}
\usepackage{float}

\begin{document}

\title{DeID-GPT: Zero-shot Medical Text De-Identification by GPT-4}

\author{Zhengliang Liu$^1$, Yue Huang$^2$, Xiaowei Yu$^3$, Lu Zhang$^3$, Zihao Wu$^1$, Chao Cao$^3$\\Haixing Dai$^1$, Lin Zhao$^1$, Yiwei Li$^1$, Peng Shu$^1$, Fang Zeng$^4$, Zeyu Zhang$^3$,\\Lichao Sun$^2$, Wei Liu$^5$, Dinggang Shen$^{6,7,8}$,\\Quanzheng Li$^4$, Tianming Liu$^1$, Dajiang Zhu$^3$, Xiang Li$^4$ }

\institutes{
    $^1$ The University of Georgia, Athens, GA; $^2$ Lehigh University, Bethlehem, PA; $^3$ The University of Texas at Arlington, Arlington, TX; $^4$ Massachusetts General Hospital and Harvard Medical School, Boston, MA; $^5$ Mayo Clinic, Phoenix, AZ; $^6$ ShanghaiTech University, Shanghai, China; $^7$ Shanghai United Imaging Intelligence Co., Ltd., Shanghai, China; $^8$ Shanghai Clinical Research and Trial Center, Shanghai, China}

\maketitle

\section*{Abstract}

\textit{The digitization of healthcare has facilitated the sharing and re-using of medical data but has also raised concerns about confidentiality and privacy. HIPAA (Health Insurance Portability and Accountability Act) mandates removing re-identifying information before the dissemination of medical records. Thus, effective and efficient solutions for de-identifying medical data, especially those in free-text forms, are highly needed. While various computer-assisted de-identification methods, including both rule-based and learning-based, have been developed and used in prior practice, such solutions still lack generalizability or need to be fine-tuned according to different scenarios, significantly imposing restrictions in wider use. The advancement of large language models (LLM), such as ChatGPT and GPT-4, have shown great potential in processing text data in the medical domain with zero-shot in-context learning, especially in the task of privacy protection, as these models can identify confidential information by their powerful named entity recognition (NER) capability. In this work, we developed a novel GPT4-enabled de-identification framework (``DeID-GPT") to automatically identify and remove the identifying information. Compared to existing commonly used medical text data de-identification methods, our developed DeID-GPT showed the highest accuracy and remarkable reliability in masking private information from the unstructured medical text while preserving the original structure and meaning of the text. This study is one of the earliest to utilize ChatGPT and GPT-4 for medical text data processing and de-identification, which provides insights for further research and solution development on the use of LLMs such as ChatGPT/GPT-4 in healthcare. Codes and benchmarking data information are available at https://github.com/yhydhx/ChatGPT-API.}

\section*{I. Introduction}

The widespread digitization of medical data has revolutionized healthcare by enabling the easy and efficient sharing of patient information \cite{Roberto23} \cite{mcdowell2022oral}. The corresponding electronic health record (EHR) systems offer a promising repository of data that can be utilized to expedite the implementation of data-driven solutions and research. At the same time, it has also raised concerns regarding the privacy and security of sensitive medical information \cite{Tang23} \cite{Paul23}. For example, clinical notes, including physician consultation, nursing assessments, discharge reports, procedure and operative reports, and radiology/pathology reports, are typically archived in a free-text format that frequently incorporates identifiable or confidential patient information. As such, unauthorized access to this information can pose significant risks to patients' confidential information and privacy \cite{urbain2022natural} \cite{tayefi2021challenges}. Patient privacy is always the top concern when sharing, uploading, and processing health information. The U.S. Health Insurance Portability and Accountability Act (HIPAA) mandates the removal of 18 categories of re-identifying information from medical records before their dissemination to preserve the confidentiality of patients \cite{Tanbir20}. It has been an urgent and essential topic for researchers to study potential ways to mitigate related concerns to apply data masking techniques to conceal sensitive data from unauthorized access \cite{Nandita21} \cite{Asokan19}.

Recently, large language models (LLM), such as OpenAI's ChatGPT and GPT-4, have shown tremendous potential in analyzing and processing textual data, thus providing opportunities for downstream tasks in medical data analysis \cite{thoppilan2022lamda,dai2023chataug,liu2023artificial,li2023artificial1,liu2023radonc,liu2023evaluating,liu2023summary,holmes2023evaluating}. LLMs such as ChatGPT can engage in conversation and generate contextual responses in a more naturalistic manner \cite{Long22}. The latest released GPT-4 can generate, edit, and collaborate with users on creative and technical writing tasks\cite{zhong2023chatradio,liu2023tailoring,liu2023radiology,liu2023radiology1}, as well as interpret and generate images \cite{liu2023holistic,yan2023multimodal}. New interesting findings have revealed that LLMs demonstrate a notable aptitude for in-context zero-shot and few-shot learning \cite{Takeshi22,Monica22,Hongjin22,shi2023mededit,guan2023cohortgpt}. This discovery boosts the development of the `prompt engineering' technique, which involves providing the LLMs with a brief contextual cue to aid in addressing the given tasks. Due to their ability to generate coherent and contextually appropriate responses, LLMs have been used to produce clinical notes and radiology reports \cite{jeblick2022chatgpt,Som23,ma2023impressiongpt}. Surprisingly, ChatGPT and GPT-4 also reveal their potential to mask sensitive or private information in medical data while preserving the overall structure and meaning of the text. Given the promising advantages of the LLMs, the objective of the work is to streamline research on clinical notes, particularly narrative free-text notes such as physician and nursing notes, by de-identifying large-scale patients' sensitive information accurately and efficiently, without manual intervention before medical data sharing and re-using to meet the HIPAA requirements. \cite{Carmel22}. 

The LLM-based methods have several potential advantages in privacy protection. First of all, LLMs have better accuracy in identifying confidential information. Large language models can leverage their ability to learn from vast amounts of data to identify patterns and relationships between words, phrases, and other elements of the text. This makes them well-suited for de-identification tasks, where the goal is to remove identifying information while preserving the meaning of the text. For example, a large language model can identify patterns in names, addresses, phone numbers, and other sensitive information and automatically remove them from text data \cite{Chengwei23}. Second, LLMs can process text data at high speed, making them an efficient tool for de-identification tasks. This is especially important when dealing with large datasets, where manual de-identification can be time-consuming and error-prone. By using a large language model, organizations, including healthcare systems, can quickly process large amounts of text data, de-identify medical records, and reduce the risk of exposing sensitive information while also complying with data privacy regulations such as HIPAA\cite{Fares23} \cite{Maad23}. Third, LLMs can be trained on different types of text data and can learn from a broad range of text data, which enables them to generalize to new, unseen data more effectively, allowing them to adapt to different de-identification tasks and use cases. By training the model on a diverse range of data, it can learn to recognize and remove a wide variety of identifying information. For example, a model trained on medical records can be used to de-identify patient information, while a model trained on financial documents can be used to de-identify credit card numbers and other financial information \cite{Xiang23}.

\begin{figure}
\begin{center}
\includegraphics[width=1.0\textwidth]{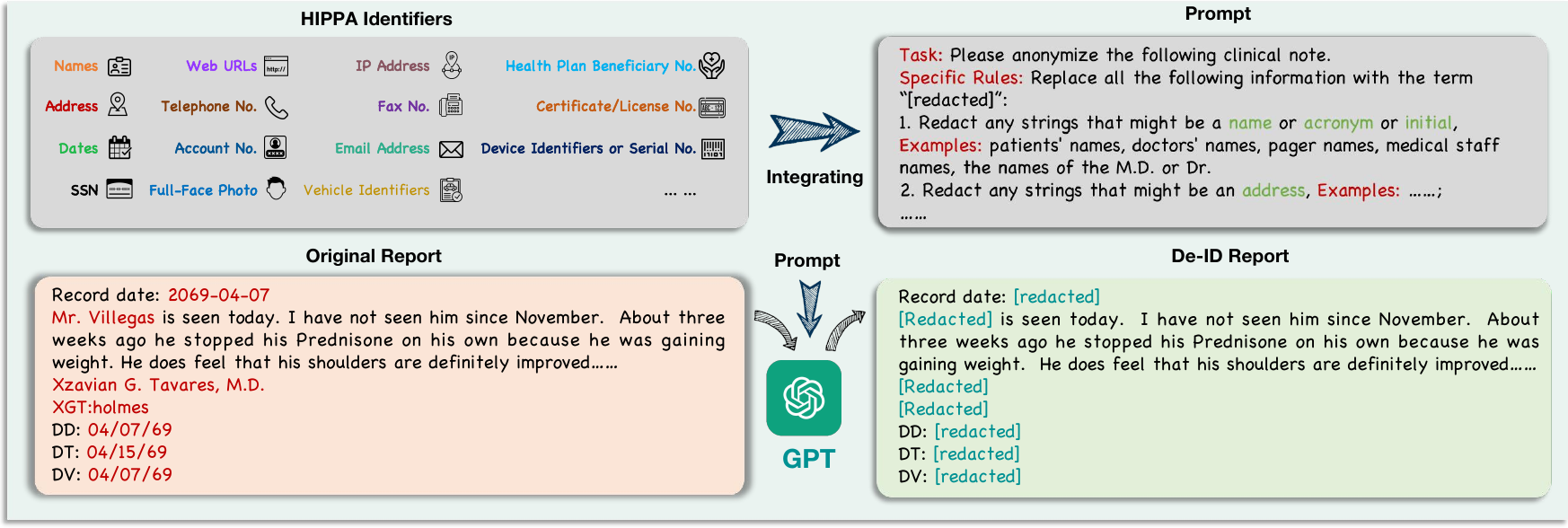}
\end{center}
\caption{The framework of the proposed DeID-GPT. DeID-GPT redacts protected health information in clinical reports through two major steps. First, HIPPA guidelines are incorporated into the designed prompts; Second, the generated prompts and the original clinical reports are sent to the LLM, such as ChatGPT and GPT-4, to generate de-identified reports.} 
\label{main}
\end{figure}

 However, LLM-based methods are still in their early stages and require further development to handle healthcare data privacy and security. To the best of our knowledge, this work is the first attempt to explore the potential of LLM for the de-identification and anonymization of medical reports. Specifically, we introduced the ChatGPT/GPT-4 into the medical domain and explored the potential of using ChatGPT/GPT-4 for data de-identification and anonymization in the medical report. Our proposed DeID-GPT framework, as illustrated in Figure \ref{main}, involves two major steps for deleting identification information. Firstly, we integrated the HIPPA Identifiers into the prompt – a set of instructions that define and customize the tasks and capabilities of LLMs. Secondly, the generated
prompts, as well as the original clinical reports, are sent into ChatGPT/GPT-4. Guided by the prompt, the model will delete the identification information in the original clinical reports. We comprehensively evaluated and compared the performance of different LLMs in de-identification tasks and provided insightful conclusions. In addition, we provide a comprehensive review of data masking techniques and their limitations and demonstrate how ChatGPT/GPT-4 can be used to overcome these limitations. We proposed potential solutions to address these concerns. It is noted that our paper itself does not use any original medical data but uses synthesized public medical datasets in which private information has been filtered. The key contribution of our work is to design appropriate high-quality prompts to make the model efficient and effective in privacy protection.

\section*{II. Related Work}

\subsection*{Large Language Models (LLMs)}
Recently, LLMs in the natural language processing (NLP) domain have been gaining significant attention from the general public. One notable example is ChatGPT, a large language model developed by OpenAI that has been a game-changer in the field of conversational AI, allowing for more natural and human-like interactions between machines and humans. The impressive achievements of ChatGPT have fueled the belief that we are entering a new era of AI, with potentially significant impacts on our society as a whole. By reflecting on the development history of language models, we can gain valuable insights into the evolution of NLP and the ongoing challenges and opportunities in the field. It is crucial to continue to innovate and push the boundaries of what is possible, while also being mindful of the ethical considerations that come with the increasing sophistication of these models. Traditional methods for generating sentences involved learning word distribution using N-gram language modeling \cite{bengio2000neural}, followed by a search for the best sequence. However, this method was limited in its ability to adapt to long sentences. To address this problem, recurrent neural networks (RNNs) \cite{mikolov2010recurrent} were introduced for language modeling tasks, allowing for the modeling of relatively long sentences. This was followed by the development of Long Short-Term Memory (LSTM) \cite{graves2012long} and Gated Recurrent Unit (GRU) \cite{dey2017gate}, which leveraged gating mechanisms to control memory during training. These approaches could attend to around 200 tokens in a sample \cite{khandelwal2018sharp}, representing significant improvement over N-gram language models.

In 2017, the Transformer model was introduced by Vaswani et al. for NLP tasks and has since become the backbone of many LLMs \cite{devlin2018bert,liu2019roberta,brown2020language, radford2018improving, radford2019language,zhang2022opt}. The Transformer model was developed to overcome the limitations of traditional models such as RNNs in handling variable-length sequences and context awareness. It consists of an encoder and an decoder, with the former taking in the input sequence and generating hidden representations, and the latter taking in the hidden representations and producing an output sequence. Each layer of the encoder and decoder contains a multi-head attention and a feed-forward neural network. The multi-head attention is the central component of the Transformer and allows the model to assign different weights to tokens based on their relevance, which helps to improve its performance in handling long-term dependencies in a wide range of NLP tasks. Another advantage of the Transformer is its highly parallelizable architecture, which enables large-scale pre-training and makes it adaptable to various downstream tasks while allowing data to trump inductive biases \cite{elhage2021mathematical}.

Since its introduction, the Transformer architecture has become the dominant choice in NLP due to its parallelism and learning capabilities. Transformer-based pre-trained language models can be broadly classified into two types based on their training tasks: autoregressive language modeling (decoders) and masked language modeling (encoders) \cite{qiu2020pre}. Masked language modeling, such as BERT \cite{devlin2018bert} and RoBERTa \cite{liu2019roberta}, involves predicting the probability of a masked token given contextual information. In contrast, autoregressive language modeling, such as the GPT family \cite{brown2020language, radford2018improving, radford2019language} and OPT \cite{zhang2022opt}, is focused on modeling the probability of the next token given the previous tokens, and is better suited for generative tasks. The GPT model, a transformer-based autoregressive decoder model that uses self-attention mechanisms to process all words in a sequence simultaneously, is one of the most notable examples of autoregressive language modeling. GPT is trained on a next word prediction task based on previous tokens, allowing it to generate coherent text. Subsequently, GPT-2 \cite{radford2019language} and GPT-3 \cite{brown2020language} continue to use the autoregressive left-to-right training method, while scaling up the number of model parameters and incorporating diverse datasets beyond basic web text, achieving state-of-the-art results on numerous NLP tasks. In addition to the two types of models based on the encoder or decoder architecture of Transformer, there are several LLMs that employ a complete encoder-decoder structure. Text-to-Text Transfer Transformer (T5) \cite{raffel2020exploring} and BART \cite{lewis2019bart} are two examples of such models. T5 is one of the primary encoder-decoder methods that utilizes a "text-to-text" approach. This means that both input and output data are transformed into a standardized text format, enabling T5 to be trained on various NLP tasks, including machine translation, question-answering, and summarization, using the same model architecture. Another frequently used encoder-decoder method is BART, which combines the bidirectional encoder from BERT and the autoregressive decoder from GPT. BART takes advantage of the bidirectional modeling abilities of the encoder while retaining the autoregressive properties for generative tasks.

Recently, a number of very large language models have emerged, including Megatron-Turing Natural Language Generation (MT-NLG), a monolithic transformer English language model with 530 billion parameters that outperforms prior state-of-the-art models in zero-, one-, and few-shot settings \cite{Smith2022}. Pathways Language Model (PaLM), a dense decoder-only Transformer model, has 540 billion parameters and is trained with the pathways system. It achieved state-of-the-art few-shot performance across most tasks, with significant margins in many cases \cite{Chowdhery2022PaLM}. Additionally, BigScience Large Open-science Open-access Multilingual Language Model (BLOOM) is another transformer-based large language model, which is trained on approximately 176 billion parameters \cite{Laurencon2023}. Furthermore, Jurassic-1 models come in two sizes, with the Jumbo version (178 billion parameters) being the largest and most sophisticated language model ever released for general use by developers \cite{Lieber2021}. 

\subsection*{Zero-shot and Few-Shot In-Context Learning}
Recently, with the development of LLMs like GPT-3 and GPT-4, which are pretrained on massive datasets and capable of capturing a wide range of tasks and knowledge, zero-shot and few-shot in-context learning have become feasible, pushing LLMs into real-world applications. This is because these models leverage prior knowledge obtained from pretraining on diverse tasks, allowing them to quickly adapt to new tasks without the need for collecting labeled data for extensive fine-tuning, which can be particularly challenging in medical fields where labeled data are limited or not available at all \cite{zhang2023biomedgpt}.

Specifically, for LLMs in NLP, zero-shot and few-shot in-context learning refer to the model's ability to understand and perform a new task by simply providing a few examples of the desired input-output pairs ~\cite{dai2023chataug} within the prompt, or even just the task instructions without any examples. The prompts help the model grasp the structures and patterns of the task, while zero-shot and few-shot in-context learning behave similarly to explicit fine-tuning at the prediction level, the representation level and the attention behavior level, enabling it to generalize and perform the new task even better without further training or fine-tuning \cite{dai2022can}, and reducing the possibility of overfitting on downstream labeled training data. While no fine-tuning is needed for these LLMs, the trade-offs include increased computational costs during inference and the potential need for expert knowledge to craft effective prompts with examples.

\subsection*{Prompt Engineering}
LLMs are a promising tool in domains where humans and AI work together to create software-reliant systems more quickly and reliably \cite{carleton2022architecting}. However, the process of collecting and labeling responses for training or fine-tuning NLP models is time-consuming and costly. Recent studies suggest that large-scale pre-trained language models (PLMs) can be adapted to downstream tasks without fine-tuning by using prompts \cite{liu2023pre}.

A prompt is a set of instructions that customizes an LLM's capabilities and influences subsequent interactions and outputs. Prompts can do more than just filter information or dictate output types; they can be engineered to create entirely new interaction paradigms, like generating quizzes or simulating a Linux terminal window \cite{white2023prompt}. Additionally, prompts have the potential to suggest other prompts for self-adaptation, making them a valuable tool in NLP \cite{white2023prompt}. In general, prompt engineering leads to a new paradigm in NLP. The advanced capabilities of prompts highlight the importance of engineering them to provide values beyond simple text or code generation. However, finding the most appropriate prompt presents a new challenge. Currently, prompts can be either created manually \cite{liu2021gpt,schick2020s} or learned  automatically \cite{liu2023pre, shin2020autoprompt}. While automatically learned prompts can yield better performance in some tasks, they are often not human-readable. Therefore, in domains where interpretability is essential, such as medical domains, manually created prompts are more commonly used.

Although prompt engineering is still at its early stage, some valuable insights into effective prompt patterns have been proposed \cite{white2023prompt}. For example, researchers in \cite{white2023prompt} compared prompt patterns to well-known software patterns, as both offer reusable solutions to common issues in a specific context, such as output generation and interaction when working with LLMs. They described 16 prompt patterns that have proven effective in improving the outputs of LLM-based conversations. These patterns have provided a great deal of inspiration, and as the field advances, it is likely that even more effective prompt patterns will be discovered.

\begin{figure*}[htp]
\begin{center}
\includegraphics[width=1.0\textwidth]{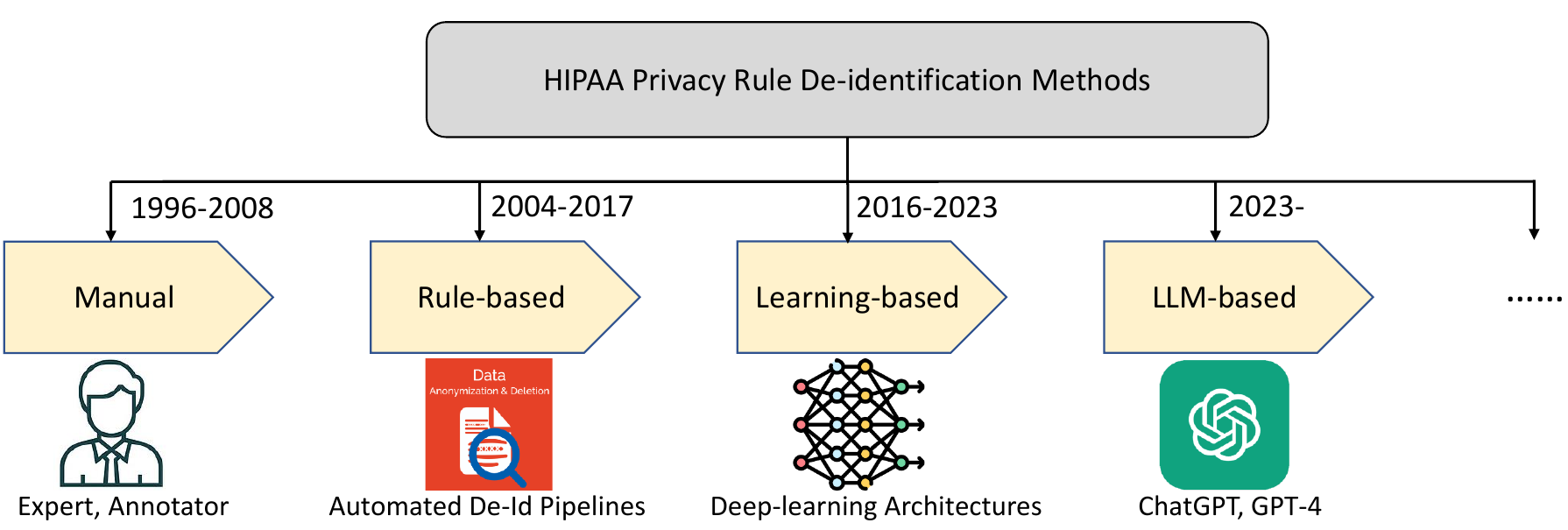}
\end{center}
\caption{Development history of de-identification methods in accordance with HIPAA.} 
\label{deidMethods}
\end{figure*}

\subsection*{De-Identification Methods}
Significant progresses have been made in filtering out confidential information from medical data, but they are usually time-consuming and far from satisfactory in practical use \cite{Thomas22} \cite{Iyadh22}. The existing de-identification methods for medical text data are mainly manual, rule-based, and learning-based. Figure \ref{deidMethods} shows the brief development of de-identification methods.

\textbf{Manual Methods} The manual de-identification of EHRs involves human annotators who read and remove sensitive information. However, this approach is both costly and time-consuming. The existing report shows that the cost of using human annotators is \$50 per hour, and they can read around 20,000 words per hour \cite{Cliffford05}. For a dataset containing 100 million words, the cost would be a staggering number of \$250,000. Additionally, the recall value for manual de-identification ranges from 63\% to 94\%, depending on the annotators \cite{Ishna08}. Besides, the manual de-identification process is prone to errors, making it an unreliable and inefficient approach. 

\textbf{Rule-based Methods} Considering the limitations of manual de-identification, automated approaches have emerged as a logical advancement for automatically de-identifying EHRs. These automated de-identification approaches are predominantly rule-based. Rule-based methods mainly depend on pre-defined word patterns with regular expressions and look-up searching dictionaries. For example, pseudonymisation \cite{Hercules19} aims to de-identify clinical data by either removing entire sentences that contain sensitive information or by substituting sensitive words with realistic alternatives, which may undermine the performance of the model \cite{Hanna20}. A hybrid model was proposed that utilizes machine learning techniques in conjunction with keyword-based and rule-based approaches, incorporating a diverse array of linguistic features, task-specific features, and regular expression template patterns to effectively address the complexities inherent in personal health information categorization \cite{Hui15Automatric}. While these systems are relatively simple to create, they lack generalizability due to the need for fine-tuning the rules for each dataset, and they fail to consider the contextual nuances of words. 

\textbf{Learning-based Methods} Learning-based methods built scalable pipelines for de-identification using machine learning and deep learning methods \cite{Joffrey20} \cite{Junhak22}, however, this usually results in unstable performance when the real data has a domain shift from the training data \cite{Urbain22} \cite{Xi19}. Though deep-learning-based methods with advanced techniques from the NLP community have been proposed to improve de-identification ability on cross-institute datasets, these methods are only trained and tested on rather small datasets and depend heavily on pre-preprocessing pipelines, such as pre-processing clinical notes \cite{Xi19}. During the COVID-19 pandemic, cross-lingual transfer learning was applied to de-identify medical records written in a low-resource language using the models built with the data written in high resources languages \cite{catelli2020crosslingual}. Improved named entity recognition (NER) model was proposed that incorporated a multi-faceted input embedding layer, consisting of standard word embedding, context-based word embedding, character-level word embedding via a convolutional neural network (CNN), external knowledge sources, and one-hot vector word features \cite{syed2022deidner}.

\begin{figure*}[htp]
\begin{center}
\includegraphics[width=1.0\textwidth]{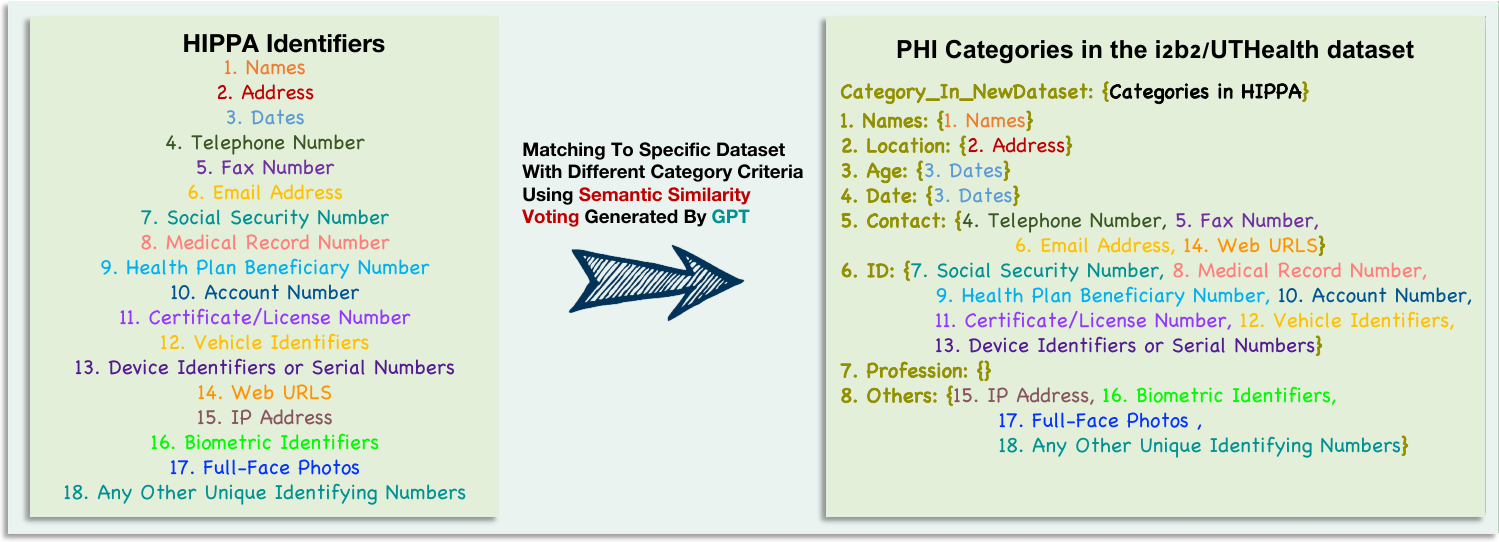}
\end{center}
\caption{To match HIPAA identifiers to dataset-specific Protected Health Information (PHI) categorization, we use a semantic similarity voting approach. For each HIPAA identifier, we calculate its semantic similarity with every category in the dataset-specific PHI category. If the similarity value exceeds a predefined threshold, we assign the identifier to the category with the greatest similarity. However, if the greatest similarity falls below the threshold, we assign the identifier to an "Others" category. This approach ensures that the HIPAA identifiers are accurately mapped to the categories while also accounting for cases where no sufficiently similar category is found. To ensure both accuracy and consistency, the semantic similarity was calculated by the same GPT used to delete the identification in later steps. This approach allowed us to leverage the GPT-4's language processing capabilities to accurately match identifiers to categories while also guaranteeing that the new categorization would be easily understood by the GPT-4 model.} 
\label{matchin} 
\end{figure*}


\subsection*{Applications of ChatGPT/GPT-4}
ChatGPT/GPT-4 presents an exciting opportunity to advance NLP and natural language understanding (NLU) across various applications. It has the potential to power chatbots, virtual assistants, and other conversational interfaces, which are becoming increasingly important as more people rely on voice and text to interact with technology. In this section, we comprehensively reviewed the applications of ChatGPT/GPT-4 in various domains.

\textbf{Healthcare:}
A prominent application prospect of LLMs is LLM-powered healthcare. Clinical practices generate exceedingly large amounts of texts on a daily basis. It is beneficial to unlock the potential of clinical text data through state-of-the-art tools and methods. The inception of the Transformer ~\cite{vaswani2017attention} has led to the rise of pre-trained language models in healthcare ~\cite{wu2022survey,liu2022survey}. For example, the ClinicalRadioBERT is a specialized language model for radiation oncology ~\cite{rezayi2022clinicalradiobert}. Other applications include radiology report summarization ~\cite{cai2021chestxraybert}, mental disorder detection ~\cite{ji2021mentalbert}, COVID-19 research summarization ~\cite{cai2022covidsum}, and clinical information extraction ~\cite{agrawal2022large}. However, the scale of existing language models is not comparable to recent developments such as ChatGPT and GPT-4. Indeed, larger models have learned from more materials and typically have more capacity to handle diverse and complex tasks. Therefore, we foresee broad adoption of ChatGPT and GPT-4 in a wide range of tasks, ranging from medical triage and question answering for patients, to knowledge-guided physicians assistance tools and fully-automated information extraction and collection. In addition, the recent popularity of LLMs is also pushing the boundary of language models, because the widespread success might break barriers that have prevented in-depth adoption of LLM-based methods ~\cite{abdullah2023chatgpt,haleem2023era}. The next generation of ChatGPT will be powered by GPT-4, which is a considerable upgrade from GPT-3.5, and we believe this will bring more innovations to healthcare NLP. Finally, it should be noted that ChatGPT is trained with RLHF. This process incorporates human preference, human values and guardrails (e.g., the exclusion of toxic generation results) into this LLM. This capability distinguishes ChatGPT from other LLMs, especially since it is extremely valuable for LLMs to understand human values and patient-centered guidelines.   

\textbf{Social Media:}
ChatGPT/GPT-4 can be used in a variety of ways on social media platforms. One way is through chatbots, which use NLP to interact with users and provide personalized responses. Chatbots powered by ChatGPT can provide more sophisticated and nuanced responses than traditional chatbots, as they can understand the contexts and intents behind user queries. ChatGPT can also be used to analyze social media data, including posts, comments, and tweets. This can provide valuable insights into user behavior and sentiment, which can help businesses and organizations make more informed decisions. For example, ChatGPT/GPT-4 can analyze social media data to identify patterns and trends, such as which products are the most popular or which topics are generating the most discussion. Another application of ChatGPT on social media is content creation. ChatGPT/GPT-4 can generate text-based contents, including blog posts, social media updates, and even news articles. This can be particularly useful for businesses and organizations that need to produce a large amount of contents quickly. ChatGPT-generated contents can also be personalized and tailored to specific audiences, which can help increase engagement and drive internet traffics to social media pages. Finally, ChatGPT/GPT-4 can be used for social media monitoring and moderation. ChatGPT/GPT-4 can analyze social media content in real-time to identify potentially harmful or inappropriate contents, such as hate speech or cyberbullying. This can help social media platforms maintain a safe and positive environment for users. Overall, the use of ChatGPT/GPT-4 on social media can help businesses and organizations improve engagement, generate high-quality content, and maintain a safe and positive environment for users.

\textbf{Content Generation:} As an AI chatbot, ChatGPT/GPT-4 can automatically produce high-quality, naturally-sounding contents to save writers' time and efforts. ChatGPT/GPT-4 is a powerful tool for generating text-based content, such as articles, blog posts, social media posts, and product descriptions. It works by training on large datasets of text and then using that knowledge to generate new text that is similar in style and content. One of the key benefits of using ChatGPT/GPT-4 for content generation is its ability to produce highly personalized content. ChatGPT/GPT-4 can be trained on a specific topic or audience, and can then generate content that is tailored to that audience. This can help businesses and organizations produce highly targeted content that resonates with their target market. Another advantage of using ChatGPT/GPT-4 for content generation is its ability to generate large volumes of content quickly. This can be particularly useful for businesses and organizations that need to produce a high volume of content, such as e-commerce sites or news outlets. By using ChatGPT/GPT-4, these organizations can quickly generate high-quality content that meets their needs. ChatGPT/GPT-4 can also be used to optimize existing content. Taking academic writing as an example, ChatGPT/GPT-4 can provide suggestions for improving writing, such as identifying grammatical errors and offering alternative phrasing options. This can be particularly useful for non-native English speakers who may struggle with writing academic papers in English. However, there are some limitations to using ChatGPT/GPT-4 for content generation. While ChatGPT/GPT-4 is able to generate high-quality content, it is still not able to produce content that is as nuanced or creative as a human writer. Additionally, ChatGPT/GPT-4 may produce content that is biased or inaccurate, as it is only as good as the data it is trained on. As a result, it is important to carefully review and edit content generated by ChatGPT/GPT-4 before publishing it. Overall, ChatGPT/GPT-4 is a powerful tool for content generation that can help businesses and organizations produce high-quality, personalized content quickly and efficiently.

\textbf{Search Engine:} 
ChatGPT/GPT-4 has the potential to revolutionize the way we approach search engines. Unlike traditional search engines, which rely on rigid keyword matching and indexing, ChatGPT/GPT-4 offers a more dynamic and personalized approach to search. By leveraging its natural language processing capabilities, ChatGPT/GPT-4 can provide more conversational and contextual responses that are tailored to the user's specific needs. One of the key advantages of ChatGPT/GPT-4 as a search engine is its ability to understand and respond to natural language queries. This means that users can ask questions in the same way they would ask a person, rather than having to rely on specific keywords or phrases. This makes the search process more intuitive and user-friendly and can help to reduce the frustration and complexity associated with traditional search engines. Another advantage of ChatGPT/GPT-4 is its potential application in the field of customer service. Companies can integrate ChatGPT/GPT-4 into their websites or applications to provide a more natural and conversational interface for users to ask questions and receive relevant answers. This can help to improve the overall customer experience and reduce the workload of customer service representatives. ChatGPT/GPT-4 can also be trained on specific domains or industries, making it an efficient and effective search engine for niche topics. This could be particularly useful in areas such as healthcare, finance, and law, where specialized knowledge and expertise are required. By providing more targeted and relevant results, ChatGPT/GPT-4 can help to streamline the search process and save users time and effort. Another potential use case for ChatGPT/GPT-4 as a search engine is in the educational field. Students and researchers can use ChatGPT to ask questions related to their academic pursuits, such as finding relevant research papers or definitions of technical terms. ChatGPT/GPT-4 can also provide personalized recommendations based on the user's past queries and search history, helping to guide the user towards relevant and useful resources. In conclusion, ChatGPT/GPT-4 has the potential to transform the search engine landscape by providing a more conversational, intuitive, and personalized approach to search. With its natural language processing capabilities and ability to be trained on specific domains or industries, ChatGPT/GPT-4 can be used in a variety of applications, from customer service to education, and beyond.

\textbf{Coding:}
Using ChatGPT/GPT-4 for coding has several benefits. The tool's ability to understand natural language inputs allows developers to input code snippets or commands in a more intuitive and user-friendly way. Additionally, ChatGPT/GPT-4 can provide contextually relevant information, saving developers time and providing accurate information. Furthermore, ChatGPT/GPT-4 can generate new code, making it an efficient tool for code generation. It can significantly improve productivity, especially for developers working on large projects. However, there are limitations to ChatGPT/GPT-4. Its cost and accessibility are limited, making it less accessible to small businesses and individuals. Additionally, it cannot fully understand the nuances of programming languages and is unable to handle certain types of queries, such as debugging or performance optimization. To ensure accurate and effective coding, it is important to use ChatGPT/GPT-4 in conjunction with other tools and resources. Improving the cost and accessibility of ChatGPT/GPT-4 would make it more widely available and useful for developers.

\textbf{Detect Security Vulnerabilities:}
ChatGPT/GPT-4 can be used to detect security vulnerabilities in a variety of ways, including analyzing code snippets, logs, and other text-based data. This can help identify potential vulnerabilities in software systems before they can be exploited by attackers. One way ChatGPT/GPT-4 can be used to detect security vulnerabilities is by analyzing code snippets. Developers can provide ChatGPT/GPT-4 with code snippets or specific commands, and the tool can analyze the code to identify potential security risks. For example, ChatGPT/GPT-4 can detect code that is vulnerable to SQL injection attacks or cross-site scripting attacks. It can also identify code that is not properly validating user input, which can lead to security vulnerabilities. Another way ChatGPT/GPT-4 can be used to detect security vulnerabilities is by analyzing security logs. Security logs can contain a wealth of information about potential security threats, such as failed login attempts, suspicious activity, or other unusual behavior. ChatGPT/GPT-4 can analyze these logs to identify patterns that may indicate a security breach or other potential security risk. ChatGPT/GPT-4 can also be used to analyze security policies and procedures. For example, developers can ask ChatGPT questions about security best practices or provide it with security policies to analyze. ChatGPT/GPT-4 can then identify potential gaps in security procedures and provide recommendations for improvement. Overall, the use of ChatGPT/GPT-4 in security vulnerability detection can help developers identify potential security risks more efficiently and effectively. This can help improve the overall security of software systems and protect against potential attacks.

\section*{III. Datasets}

\subsection*{Data Description}
\begin{figure*}[htp]
\begin{center}
\includegraphics[scale=0.4]{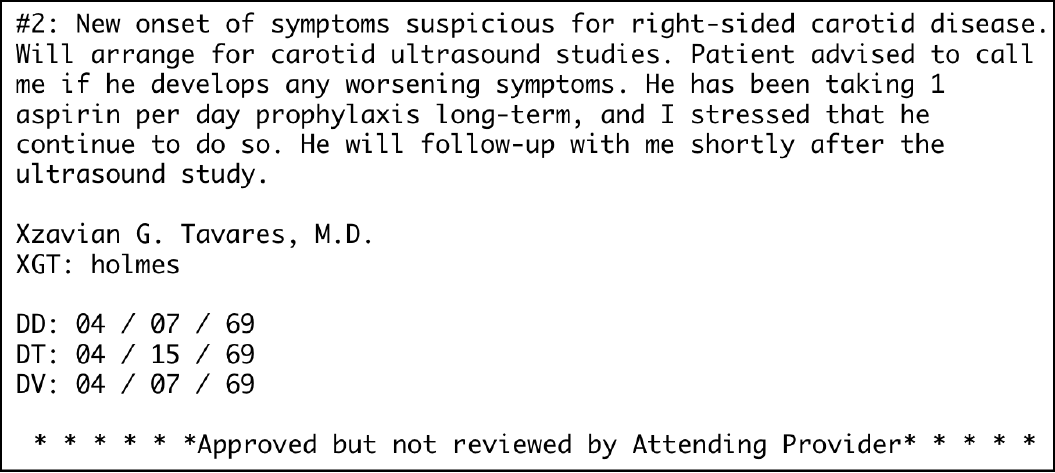}
\end{center}
\caption{Sample original clinical note from the 2014 i2b2/UTHealth dataset.} 
\label{sample_original_fig}
\end{figure*}

\begin{figure*}[htp]
\begin{center}
\includegraphics[scale=0.4]{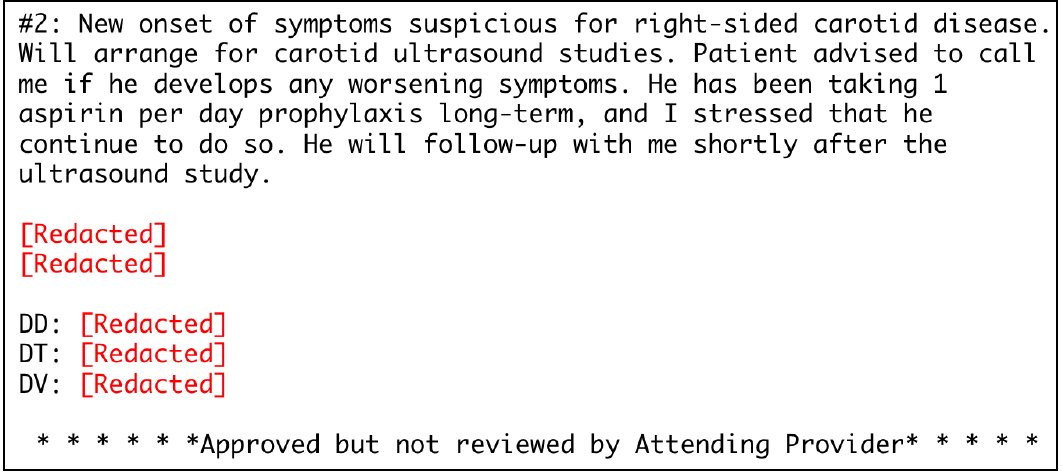}
\end{center}
\caption{Sample de-identified clinical note from the 2014 i2b2/UTHealth dataset.} 
\label{sample_redacted_fig}
\end{figure*}

\textbf{The i2b2/UTHealth Challenge:}
We benchmark our proposed method using the 2014 i2b2/UTHealth de-identification challenge dataset ~\cite{stubbs2015annotating}. Upon request, the Blavatnik Institute of Biomedical Informatics at Harvard University granted us access to this dataset. This dataset contains 1,304 free-form clinical notes of 296 diabetic patients. All PHI entities were manually annotated and replaced with surrogates. Specifically, names, professions, locations, ages, dates, contacts and IDs were replaced by surrogate information to protect privacy and facilitate de-identification research. For example, if there is a real patient named "Mr. James McCarthy" who visited the hospital on 12/01/2013, these strings will be replaced by "Mr. Joshua Howard" and "04/01/2060", respectively. Figure \ref{sample_original_fig} presents a sample clip of this data, and figure \ref{sample_redacted_fig} presents the same excerpt de-identified by ChatGPT/GPT-4.

\section*{IV. Methodology}

In this section, we will describe the methodology of this work. Our primary approach is to utilize API access and manual testing to evaluate ChatGPT (powered by GPT-3.5) and GPT-4's (through OpenAI's web interface that is shared with ChatGPT)performance on anonymizing clinical notes. We will describe our entire workflow in detail, from the data preprocessing to evaluation. 

It is noteworthy to point out that the grand paradigm shift from fine-tuning to prompt-based in-context learning has revolutionized the NLP field. In this study, we intend to devise a new strategy to employ the zero-shot capability of very recent language models to complete the data de-identification process. We carefully design prompts that work well with ChatGPT and GPT-4 to generate the best results with minimal human annotation efforts. Thanks to the scale of LLMs and the power of in-context learning, the presented framework requires no change when being applied to different data. We present a full pipeline that is straightforward to implement and naturally explainable.  

\subsection*{Data Preprocessing}
The original Harvard 2014 i2b2/UTHealth de-identification challenge dataset is stored as XML files. One XML file corresponds to one complete clinical note that documents the symptoms, clinical records and medical impressions of one particular visit. Such files consist of various XML tags that correspond to different information in the clinical notes.

We have implemented in-house scripts to extract information from these XML files and store them in a reference database. The main text of the clinical notes are further cleaned and stored as text files. These files are the input to the LLMs audited in this study. In addition, the sensitive text entities identified by human annotators are extracted and reserved for performance evaluation. 

\subsection*{Accessing ChatGPT and GPT-4}
\begin{figure*}[htp]
\begin{center}
\includegraphics[width=0.9\textwidth]{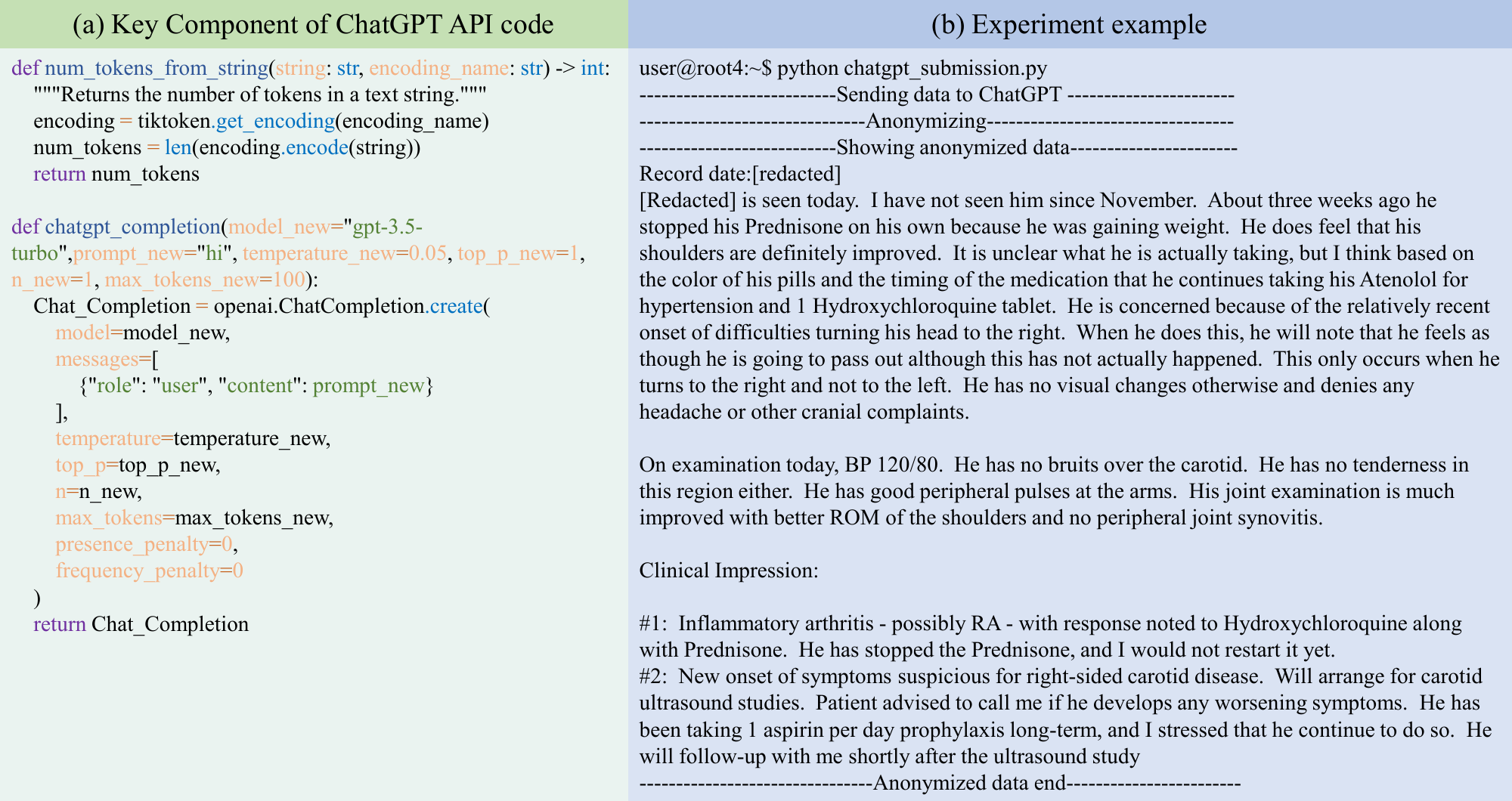}
\end{center}
\caption{The left column shows the main components of the ChatGPT API code, while the right column shows the steps involved in generating an anonymized example.} 
\label{chatgpt-code}
\end{figure*}

\begin{figure*}[htp]
\begin{center}
\includegraphics[width=0.9\textwidth]{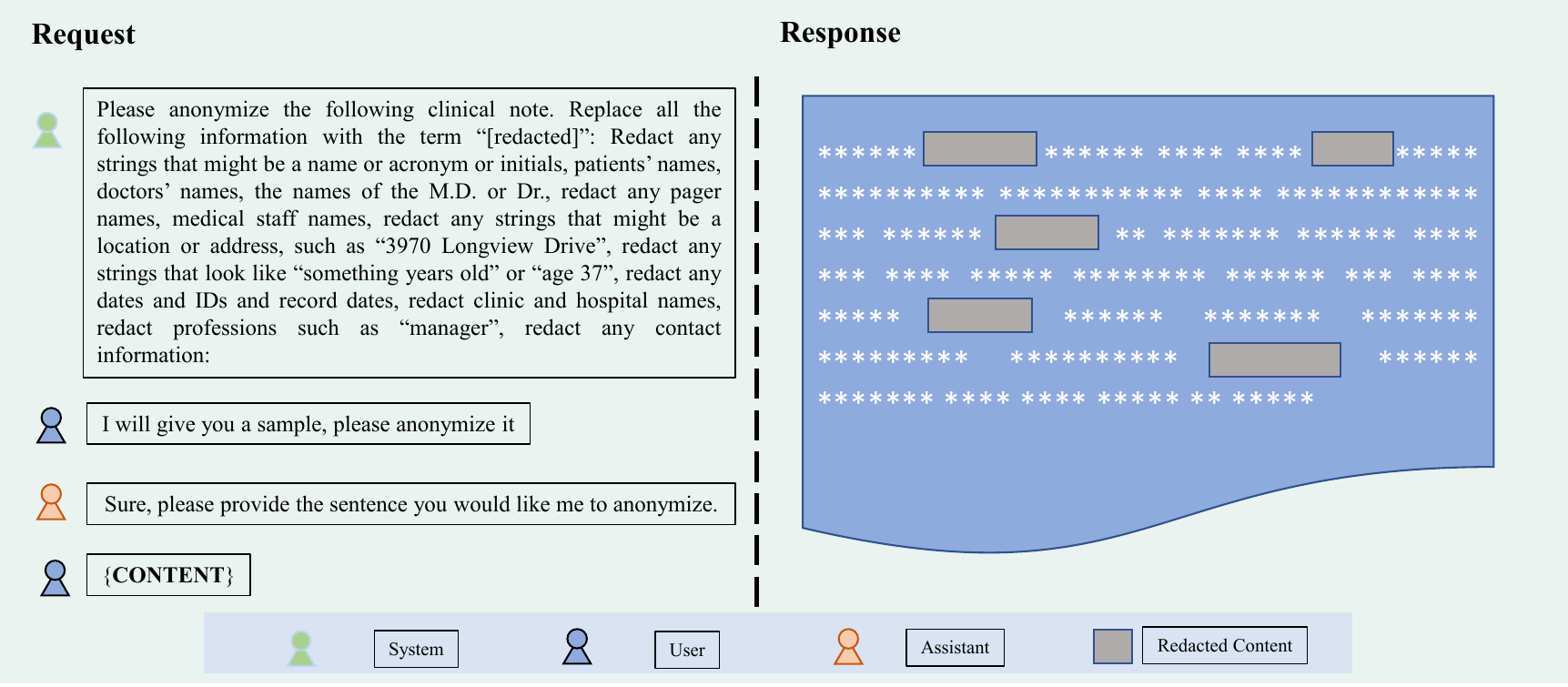}
\end{center}
\caption{To anonymize clinical notes using the ChatGPT API, we first supply prompts from the system and then send the clinical notes from the user role. The returned results from ChatGPT will be anonymized by the model, and detected sensitive PHI information will be replaced by the term 'redacted'.} 
\label{chatgpt-api}
\end{figure*}

To access the OpenAI API, people will need to create an account on OpenAI's websitrate and obtain an API key. Once you have your API key, you can use it to make API requests to OpenAI, including requests to the ChatGPT model. For more detail, please refer to our open-source code at GitHub. Fig. \ref{chatgpt-code} displays the crucial code elements of our ChatGPT API. After setting the appropriate parameters and submitting the prompts to the ChatGPT server, the generated texts can be obtained, as indicated in the right column of Fig. \ref{chatgpt-code}. The left column of Fig. \ref{chatgpt-api} depicts the step-by-step procedure for anonymizing sensitive information using our ChatGPT API. The sensitive information contained in the response text (shown in the right column of Fig. \ref{chatgpt-api}) is already concealed by ChatGPT.
Currently, there is no open access to the GPT-4 API. Therefore, we rely on manually testing on the OpenAI web interface to carry out our GPT-4 experiments. 

\begin{table}[]
\resizebox{\textwidth}{!}
{
\begin{tabular}{|l|l|l|l|}
\hline
\multicolumn{4}{|c|}{HIPAA Identifiers}  \\ \hline                                                
1   & Names   & 10 & Account numbers                                        \\ \hline
2                 & All geographical address elements smaller than state     & 11 & Certificate numbers                                    \\ \hline
3                 & All data elements related the individual (except year) & 12 & Vehicle serial numbers and identifiers                 \\ \hline
4                 & Phone numbers                                           & 13 & Device serial numbers and identifiers                  \\ \hline
5                 & Fax numbers                                             & 14 & Web resource locators (URLs) and links                 \\ \hline
6                 & Email addresses                                          & 15 & IP addresses                                           \\ \hline
7                 & Social security numbers                                 & 16 & Biometric identifiers (e.g. fingerprint)               \\ \hline
8                 & Medical record numbers                                  & 17 & Full face photographic images                          \\ \hline
9                 & Health plan beneficiary numbers                         & 18 & Any unique identifying number, code, or characteristic \\ \hline
\end{tabular}
}
\caption{List of HIPAA identifiers.} 
\label{all_hipaa_table}
\end{table}

\subsection*{A Simple and Versatile Framework}
Our goal is to develop a workflow that greatly simplifies clinical data de-identification, which can facilitate any subsequent research and collaborations. The advent of LLMs such as ChatGPT and GPT-4 has revolutionized NLP, and we see great potential in applying LLMs to the realm of privacy protection. The key advantages of LLMs can be categorized as \textbf{Simplicity}, \textbf{Annotation-free}, and \textbf{Adaptability}. These positive characteristics make LLMs the ideal tools for data anonymization applications. 

Specifically, in-context learning eliminates the need for fine-tuning, thereby saving significant amounts of time and reducing pipeline complexity ~\cite{qin2023chatgpt, zhou2023comprehensive}. Supervised fine-tuning is not a trivial task and requires sufficient experience in NLP. 

In addition, this approach has exceptional zero-shot and few-shot learning capabilities ~\cite{brown2020language, zhou2023comprehensive, bang2023multitask}, which can reduce the need for large-scale annotation efforts, as the model can learn from a relatively small amount of annotated data.

By incorporating contextual information, which allows the model to better understand the meaning and context of the data it is processing, in-context learning enables the adaptation to new tasks and domains with great ease ~\cite{brown2020language, zhou2023comprehensive}. Indeed, LLMs are very versatile and flexible. \textbf{Unlike any previous methods, no code or procedural changes are required when applying our solution to different hospitals, different languages and different data formats}. 

These advantages make in-context learning a promising approach for a wide range of NLP tasks, including de-identification. We develop this study to methodologically validate ChatGPT and GPT-4's advantages. 
\vspace{0.3cm}
\begin{algorithm}[H]
\caption{Framework of DeID-GPT for few-shot text classification.}
\label{algorithm1}
\textbf{Input}: original dataset ${D}_o$ and HIPAA-compliant prompt ${P}_h$  \\
\textbf{Initialize}: Initialized ChatGPT $model$ \\
\textbf{Definition}: ${D}_o$ is original clinical notes dataset, ${D}_d$ is the de-identified dataset and \textit DeId-GPT is the de-identification method based on ChatGPT. HIPAA-compliant prompt ${P}_h$ is the prompt to guide ChatGPT \\
    \textbf{Parameters}: In-context learning epochs of the original dataset $epoch_o$
\begin{algorithmic}
    \FOR{epoch {\bfseries in} $epoch_o$}
        \STATE train($model$, $D_o$, ${P}_h$)
    \ENDFOR
    \STATE ${D}_d$  = \textit{DeId-GPT}(${D}_o$)
\end{algorithmic}
\textbf{Output}: De-Identified dataset ${D}_d$ 
\end{algorithm}

\begin{figure*}[htp]
\begin{center}
\includegraphics[width=1.0\textwidth]{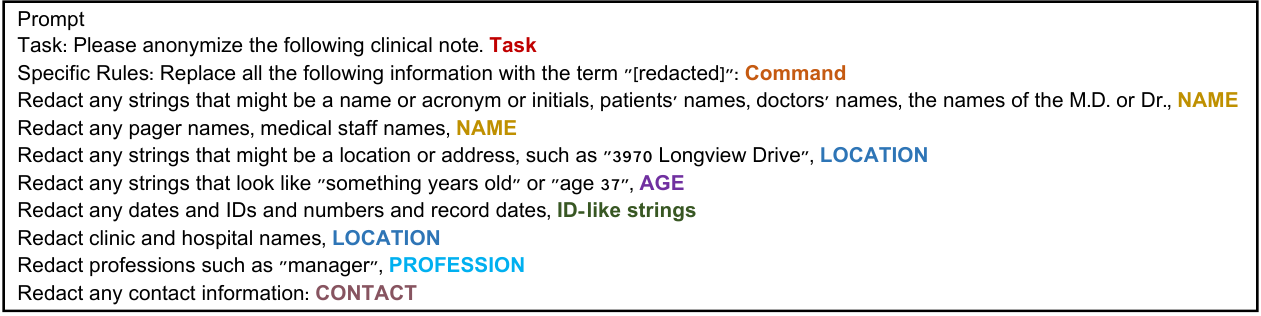}
\end{center}
\caption{This is the optimal prompt template we have designed for this task. We break down a prompt into 3 main segments: The task statement, the command, and specific rules. The task statement describes the goal and scope of the task. The command specifies actions to complete the task. Specific rules define specific requirements and optionally contain concrete examples. In the figure, we also annotate each line with a tag that describes its identity or correspondence in the PHI categories.} 
\label{fig:optimal_prompt}
\end{figure*}
\subsection*{Designing Prompts to Incorporate HIPAA Guidelines} \label{hipaa-prompt}
The HIPAA PHI categorization is the gold standard of defining clinical privacy. There are 18 HIPAA identifiers that are required to be de-identified, since this information can be used to identify, locate, or contact individuals. This is particularly important and relevant in processes (e.g, research and clinical collaborations) that involve data-sharing and transmission of clinical text documents. Figure \ref{matchin} illustrates the mapping between HIPAA identifiers (see Table \ref{all_hipaa_table}) and the i2b2/UTHealth benchmark used in this study. This correspondence relationship to HIPAA is generalizable (with proper changes) to any de-identification datasets. 

Existing research clearly indicates that LLMs produce optimal results only when provided with the right prompts. To effectively utilize LLMs to redact sensitive information and ensure proper adherence to HIPAA guidelines, we design the following optimized prompts \textbf{template} for the benchmark dataset that unleashes the potential of LLMs. 

In this template, we first specify the \textbf{task} for the LLM to complete. It is necessary to provide the task information upfront, since it is easy for the user to comprehend and explain the task to be executed by the model. In addition, given the autoregressive nature (i.e., the next token is generated based on previous tokens) of the GPT-based models ~\cite{radford2018improving}, it is important to push this information upfront for the LLM to better understand the user's demands and generate the most relevant results.   

We then specify a special rule, (e.g., "Replace all the following information with the term “[redacted]”:") that is associated with the declared task. This segment of the prompt can be adjusted to fit the specific task. We require the model to replace sensitive information with the token "[redacted]" for better \textbf{explainability}. This step also facilitates subsequent processing, including but not limited to result evaluation, surrogate information replacement (e.g., replacing the redacted information with fake synthetic data) and data-sharing. 

Finally, we explicitly define the specific information that needs to be obscured. For example, we ask the model to "Redact any strings that might be a location or address, such as "3970 Longview Drive"". Ideally, the rules laid out in this segment should correspond to various categories of PHI in the target dataset. Based on our experience, these PHI categories are typically mapped to the HIPAA PHI guidelines, and it is necessary to cover all sensitive information so that the results are aligned with the HIPAA mandate as much as possible. Generally, it is helpful to explicitly specify such information and provide examples for better results and interpretability.

\subsection*{Prompting Caveats}

\begin{figure*}[htp]
\begin{center}
\includegraphics[width=0.95\textwidth]{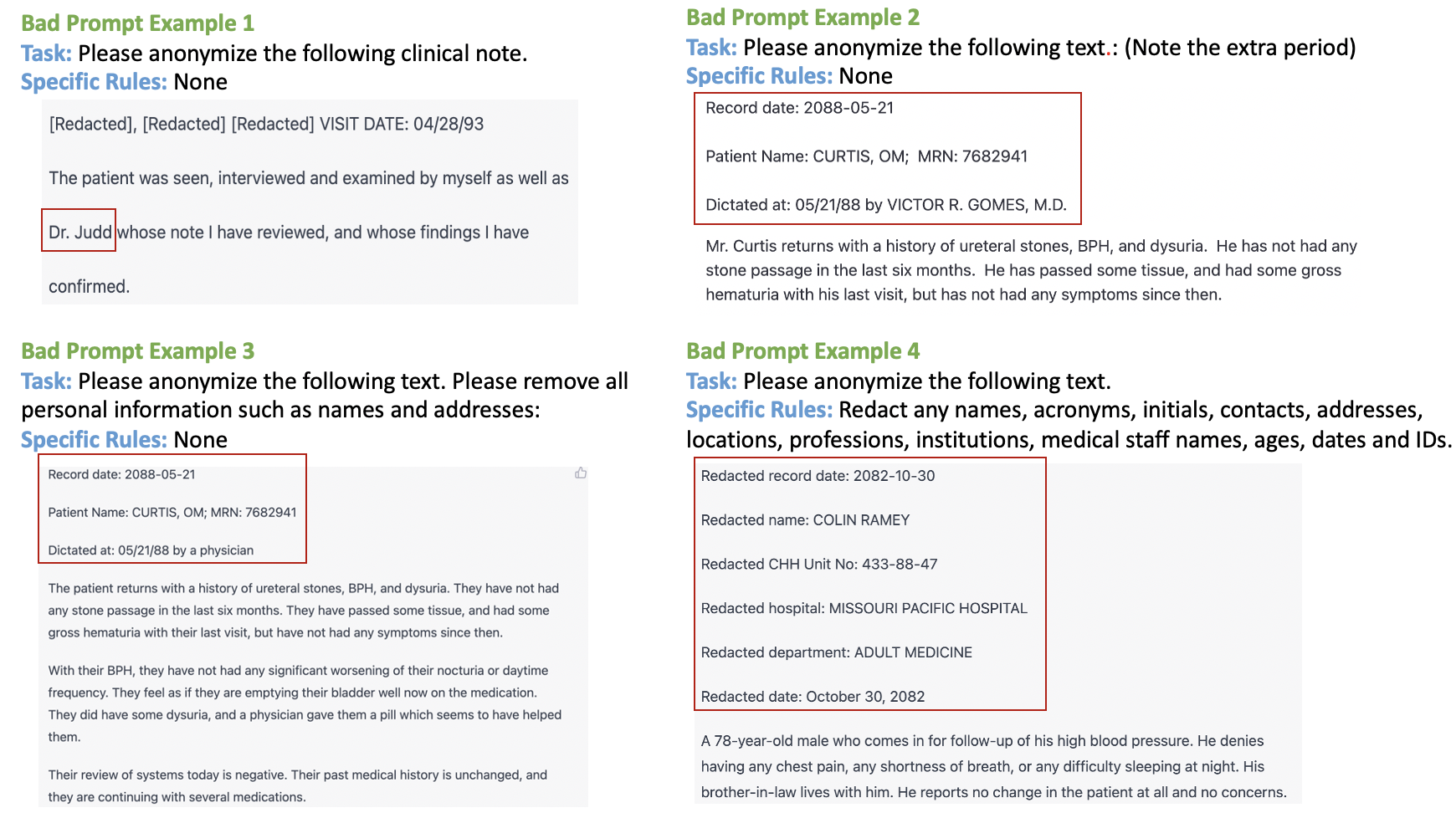}
\end{center}
\caption{We present 4 examples of bad prompts. Example1: Only stating the task in the prompt is not optimal; Example2: Punctuation matters. The extra period made the command uninterpreted by ChatGPT; Example3: Multiple-task statement confuses ChatGPT; Example4: Did not explicitly specify the desired output}
\label{faile_prompts}
\end{figure*}

Prompt design is an engineering process that combines art and science ~\cite{reynolds2021prompt,jiang2022promptmaker}. Since this is the first study on utilizing LLMs to perform de-identification and the first work in exploring the ability to de-identify clinical data, we believe it is necessary to share our experience with both good examples and counterexamples. We have presented the optimal prompt template in the previous section. In this section, we present ineffective prompts that lead to sub-optimal results. The list of ineffective prompts is non-exhaustive, and these samples are exemplary caveats.

\section*{V. Experiment \& Results}

\subsection*{Experimental Design}
We compare ChatGPT and GPT-4 against several baselines methods, BERT ~\cite{devlin2018bert}, RoBERTa ~\cite{liu2019roberta}, ClinicalBERT ~\cite{alsentzer2019publicly}. 

\textbf{BERT} is the classic transformer-based language model ~\cite{devlin2018bert}. The bi-directional transformer architecture enables impressive contextual understanding, and BERT has quickly become one of the most popular NLP models. It has been widely used in many domains and industries, and there are various offshoot models that are equipped with improved architectural design ~\cite{rogers2021primer} or are trained on domain-specific data ~\cite{gu2021domain}.  

\textbf{RoBERTa} improves from BERT by removing the next sentence prediction (NSP) objective ~\cite{liu2019roberta}. It is also trained with larger batches and on more training data. In addition, the masking patterns in RoBERTa are dynamically changed. Overall, RoBERTa can be regarded as a more refined version of the BERT model.  

\textbf{ClinicalBERT} is a BERT-based model further pre-trained on the Medical Information Mart for Intensive Care III (MIMIC-III) dataset ~\cite{johnson2016mimic}, a large collection of 2,083,180 clinical notes from the Beth Israel Deaconess Medical Center. This database contains data of 38,597 patients (admitted between 2001 and 2012) and is commonly used in clinical NLP ~\cite{pandey2022comprehensive}. The ClinicalBERT model is therefore a suitable model for medical text processing. 


For all the baseline methods, we download pre-trained weights from Hugging Face \footnote{https://huggingface.co/} and initialize them on our local servers. BERT, RoBERTa and ClinicalBERT are initialized on Nvidia 3090 GPUs with 24 GB memories. 


For all methods except GPT-4, we test all test cases in the test set of the i2b2/UTHealth data. However, since there is no public access to the GPT-4 API, one of our experts manually tested 50 random samples from the testing set to generate responses from GPT-4 using the OpenAI web interface. Regardless of the testing method, the generated responses go through the same evaluation script to calculate accuracy metrics. 


\subsection*{Results}
Experimental results show that GPT-4 achieves the highest de-identification accuracy (over 0.99) in a zero-shot scenario when provided with an optimal, explicitly specified prompt. It outperforms the GPT-3 powered ChatGPT and all other baselines. Table \ref{table2} presents the complete experimental results. 

\fbox{\begin{minipage}{\textwidth}
\textbf{Implicit Prompt}
"Please anonymize the following clinical note" is an example of an implicit prompt. The desired outcome is stated, but the prompt lacks specific instructions and examples that help the LLM to carry out the task. 

\textbf{Explicit Prompt}
are prompts that contain concrete information that helps the LLM generate desired results within a clearly defined space. It contains a well-written and executable description of the desired output, clearly defines the task, and explicitly provides concrete examples. Please refer to Figure \ref{fig:optimal_prompt} for an example of a good explicit prompt. 
\end{minipage}}

It is noteworthy to point out that the BERT-based baselines are fine-tuned on the dataset through a rigorous supervised learning named entity recognition (NER) process. Despite their strong performance (both RoBERTa and ClinicalBERT attain over 90\% accuracy rates), the fine-tuning process requires significantly more time and engineering efforts to complete, compared to the streamlined process offered by large language models such as ChatGPT and GPT-4. Our clinician experts believe this demonstrates the attractiveness of employing large language models over standard-sized models for the de-identification task, since LLMs require significantly less efforts and are accessible to health providers who have limited experience in machine learning.

In addition, we note that carefully crafted prompts can significantly improve LLM performance. For example, an optimally designed prompt improves ChatGPT performance from 0.686 to 0.929. GPT-4 is less susceptible to defects in the prompt, since it nonetheless performs well even when fed with a simple, implicit prompt. However, an optimally designed prompt propels the accuracy of GPT-4 to a new level that is readily usable even for real-world applications. 


\begin{table*}[t]
\centering
\caption{ Clinical notes de-identification results with LLMs.  }
\label{table2}
{
\begin{tabular}{llclclc}
\hline
\multirow{2}{*}{De-Identification}           & \multicolumn{2}{c}{i2b2}           \\ \cline{2-5} 
                                             & Implicit Prompt*           & Explicit Prompt*     & General Prompt* &Fine-tuning
                                             
                                             \\
                                             & (Zero-shot)           & (Zero-shot)     & (Zero-shot) &
                                             
                                             \\                                            
                                             
                                             \hline
ChatGPT                                          & 0.686         & 0.929 & N/Aw & N/A                            \\
GPT-4                                      & \textbf{0.908} & \textbf{0.99}  & N/A & N/A  \\
BERT                                          & -    & -  & -  & 0.798                            \\
RoBERTa                                          & -    & -  & -  & 0.947                            \\
ClinicalBERT                                          & -    & - & -  & 0.974                           \\
mT0                                          & 0.824    & N/A     & 0.827   & N/A                     \\
Falcon-7b                                          & 0.603    & 0.605  & 0.597   & N/A                      \\
Flan-t5-base                                      & N/A  & N/A  & 0.737   & N/A                                   \\
Llama1-7b                                          & 0.609    & 0.612  & 0.597   & N/A                            \\
Llama2-7b                                          & 0.609    & 0.612  & 0.597   & N/A                            \\
\hline
\multicolumn{4}{c}{\footnotesize *Note: The prompting based methods are essentially performing Zero-Shot learning}
\end{tabular}
}
\end{table*}

\subsection*{De-Identification Quality Evaluation}
The performance of the de-identification methods is evaluated through entity-wise accuracy, defined by the percentage of sensitive entities removed after the inputs are processed by the corresponding method. 
\begin{align}\label{eq_accuracy}
Accuracy = \frac{TP+TN}{TP+TN+FP+FN}
\end{align}
where TP, TN, FP, and FN denote numbers of true positives, true negatives, false positives, and false negatives, respectively.

\subsection*{Error Analysis}
We conduct an error analysis by examining the incorrect answers generated by various LLMs. We have identified significant differences in the nature of errors made by these models compared to GPT-4 and ChatGPT. The errors observed in GPT-4 and ChatGPT are predominantly associated with their limitations in recognizing relevant entities, which can be attributed to their proficiency and experience in this specific task. On the other hand, the errors in the Llama series and Falcon can be primarily attributed to a fundamental misunderstanding of the task itself, often resulting in nonsensical responses, such as merely repeating the content of the given prompt. Additionally, an excessively lengthy prompt can also lead to LLMs struggling to grasp its meaning. Overall, we conclude that only a small subset of these LLMs demonstrate the potential for swift practical application, and the majority of LLMs still have significant room for improvement.

\section*{VI. Discussion}

\subsection*{Locally-Deployed DeID-GPT for Hospital Use}
Despite the impressive performance of ChatGPT and GPT-4 for data de-identification, these models could only be accessed through online APIs, making them impossible to be applied in a hospital setting as patient data cannot be stored nor transmitted to a non-authorized external party. Furthermore, as the code and implementation details of ChatGPT/GPT-4 are not open to the public, we cannot fully validate their functionalities to ensure these models are HIPAA-compliant and meet the standards of the hospital's quality management system (QMS). Therefore, it would be vital to use open-source or in-house trained LLMs for the de-identification task and deploy them locally to ensure data security, privacy, integrity, and proper adherence to HIPAA guidelines. Thus, we are investigating solutions for the locally-deployed DeID-GPT. For example, we experimented with LLaMA ~\cite{touvron2023llama} in this work to preliminarily examine its performance. However, LLaMA failed to generate coherent and relevant outputs for all test cases, and is therefore not directly comparable to other tested methods. We also intend to try other open-source LLMs such as OPT ~\cite{zhang2022opt} and BLOOM ~\cite{scao2022bloom}. These models are open-sourced and were trained with data on the public domain, making it possible to reuse without licensing concerns ~\cite{touvron2023llama}. To deal with the challenge of limited computational resources at the local site, we will also explore the state-of-the-art model quantization ~\cite{dettmers2022llm}, acceleration ~\cite{zhang2020accelerating} and GPU parallelization ~\cite{narayanan2021efficient, zeng2022acctfm} techniques to streamline the deployment.
\subsection*{Developing Domain-specific LLMs}
To the best of our knowledge, there are no LLMs specifically tailored to the medicine and healthcare domain with medical text such as the clinical notes and radiology reports used in this work, let alone for more diversified medical specialties. Given the success of previous domain-specific language models such as BioBERT ~\cite{lee2020biobert} and BioGPT ~\cite{luo2022biogpt}, we believe it is promising and helpful to develop domain-specific LLMs. At the data level, it is possible to continue the pre-training process on domain-specific data. This is a proven route to success since it is well known that language models can perform better on domain-specific benchmarks and tasks when trained with relevant data and knowledge ~\cite{gu2021domain,rezayi2022agribert} as the models are exposed to domain vocabulary, jargons, terms and writing styles. At the architecture level, it is also possible to modify and optimize LLMs for specific domains. For example, it is possible to design modules that can more effectively capture sensitive HIPAA-protected information by prioritizing such information in the input ~\cite{liao2023mask}. It might also be insightful to explore the possibility of efficient passing of inputs with sparsely activated forward pass blocks ~\cite{fedus2021switch} to achieve performance or efficiency gains. 

\subsection*{Improving De-Identification with LLMs through Fine-tuning}
The upcoming public release of GPT-4's API service will enable fine-tuning of the model. Given the impressive performance of GPT-4 in the zero-shot prompting setting, we expect that it can deliver even better performance with fine-tuning to the domain-specific data (e.g., clinical notes and reports). On the other hand, to better ensure data privacy and security, it is also necessary to investigate the potential of replicating GPT-4-like performance locally with other LLMs. On local servers, it is more necessary and beneficial to evaluate the impact of fine-tuning on task performance, as locally-deployable models generally have a smaller parameter size. Regardless of the approaches, we see fine-tuning LLMs as a promising research direction that enables the practical use of LLMs in a hospital setting. 

\subsection*{Applying LLM to Anonymize Other Data}
Text de-identification and data anonymization are crucial in many other scenarios where personal information needs to be collected or shared and privacy protection is paramount. While our approach focuses on medical data, which requires particularly stringent privacy protections, it could be extended to other sensitive data domains beyond healthcare. For example, financial institutions, such as banks and insurance providers, may need to de-identify data containing customer information, such as account numbers and transaction details, to safeguard individuals' privacy. Market research companies also collect data on individuals' opinions, behaviors, and preferences, which should be de-identified to protect the privacy of research participants. Additionally, sensitive information may need to be shared in legal proceedings, including financial records, medical records, and personal information that needs to be de-identified to preserve individuals' privacy. Our proposed ChatGPT/GPT-4-based approach in this study could potentially be generalized to these scenarios in the future, offering a novel approach to data de-identification by utilizing LLMs.

\section*{VII. Conclusion and Future Perspective}
As far as we know, this study is the first work in the NLP literature to investigate the possibility of employing LLMs such as ChatGPT and GPT-4 for data de-identification. In particular, we partnered with clinicians to evaluate these powerfully versatile modern models on medical text anonymization. Experimental results indicate that ChatGPT and GPT-4 have excellent abilities in de-identifying medical data compared to other LLMs. The application of LLMs to medical text data has already shown promise in providing valuable insights into various medical conditions and diseases, and there is significant potential for future integration with other modalities such as medical imaging data~\cite{wang2023chatcad}. 

With its multi-modal capabilities, GPT-4 can be explored for the integration of multimodal medical data such as radiological images, pathological images, clinical text reports, and genomics data, among others, to gain new insights for disease diagnoses, treatments, follow-up, and prognosis. By enabling cross-modality analysis using GPT-4 and similar methodologies, we can significantly enhance our understanding of different medical conditions, including various types of cancers, brain disorders, cardiovascular diseases, and many other human diseases. Ultimately, the development and application of GPT-4 related approaches for multi-modal medical data have great potential to revolutionize healthcare. 

Reciprocally, healthcare professionals can also contribute to the advancement of LLMs such as GPT-4 and its related models/methods through their domain expertise and expert feedback. For instance, highly skilled and professional medical physicians can advance the reinforcement learning by human feedback (RLHF) used in ChatGPT/GPT-4 to reinforcement learning by expert feedback (RLEF) in the medical domain. We envision that this RLEF framework can significantly speed up the pace of widely adopting GPT-4 and its future variants in the medical and healthcare field.

\makeatletter
\renewcommand{\@biblabel}[1]{\hfill #1.}
\makeatother

\bibliographystyle{vancouver}
\bibliography{amia}  

\end{document}